\documentclass[3p]{elsarticle2}

\usepackage{amsmath}
\usepackage{booktabs}
\usepackage{color}
\usepackage{multirow}
\usepackage{paralist}
\usepackage{rotating}
\usepackage{tabularx}
\usepackage{url}

\usepackage[colorlinks  = true%
           ,pdfauthor   = {{Andres Sanin, Conrad Sanderson, Brian C. Lovell}}%
           ,pdftitle    = {{Shadow Detection: A Survey and Comparative Evaluation of Recent Methods}}%
]{hyperref}

\usepackage{bm}

\graphicspath{{./}{figures/}}

\newcolumntype{q}{@{\hspace{1pt}} >{\centering\arraybackslash} m{0.1345\textwidth} @{\hspace{1pt}}}

\journal{Pattern Recognition}

\renewcommand{\baselinestretch}{1.2}\small\normalsize

\begin{document}

\renewcommand{\today}{~}

\begin{frontmatter}

\title
  {
  {\large\bf Shadow Detection: A Survey and Comparative Evaluation of Recent Methods}
  }

\author
  {
  \normalsize
  Andres Sanin, Conrad Sanderson, Brian C. Lovell
  }

\address
  {
  \normalsize
  ~\\
  NICTA, PO Box 6020, St Lucia, QLD 4067, Australia\\
  The University of Queensland, School of ITEE, QLD 4072, Australia
  ~\\
  ~\\
  ~\\
  
  \begin{itemize}
     \item {\normalfont Associated C++ source code: \href{http://arma.sourceforge.net/shadows/}{http://arma.sourceforge.net/shadows/}}
     \item {\normalfont Published as:} \\\vspace{1ex}
     ~~~{\normalfont A.~Sanin, C.~Sanderson, B.C.~Lovell.}\\
     ~~~{\normalfont Shadow Detection: A Survey and Comparative Evaluation of Recent Methods.}\\
     ~~~{\normalfont Pattern Recognition, ~Vol.~45, ~No.~4, ~pp.~1684--1695, ~2012.}\\
     ~~~{\normalfont \href{http://dx.doi.org/10.1016/j.patcog.2011.10.001}{http://10.1016/j.patcog.2011.10.001}}
  \end{itemize}
  ~\\
  
  }

\begin{abstract}

This paper presents a survey and a comparative evaluation of recent techniques for moving cast shadow detection.
We identify shadow removal as a critical step for improving object detection and tracking.
The survey covers methods published during the last decade,
and places them in a feature-based taxonomy comprised of four categories:
chromacity, physical, geometry and textures.
A selection of prominent methods across the categories is compared 
in terms of quantitative performance measures (shadow detection and discrimination rates, colour desaturation)
as well as qualitative observations.
Furthermore, we propose the use of tracking performance as an unbiased approach
for determining the practical usefulness of shadow detection methods.

The evaluation indicates that
all shadow detection approaches make different contributions and all have individual strength and weaknesses.
Out of the selected methods,
the geometry-based technique has strict assumptions and is not generalisable to various environments,
but it is a straightforward choice when the objects of interest are easy to model and their shadows have different orientation.
The chromacity based method is the fastest to implement and run,
but it is sensitive to noise and less effective in low saturated scenes.
The physical method improves upon the accuracy of the chromacity method by adapting to local shadow models,
but fails when the spectral properties of the objects are similar to that of the background.
The small-region texture based method is especially robust for pixels whose neighbourhood is textured,
but may take longer to implement and is the most computationally expensive.
The large-region texture based method produces the most accurate results,
but has a significant computational load due to its multiple processing steps.

\end{abstract}


\end{frontmatter}

\newpage

\section{Introduction}
\label{sec:introduction}

Many computer vision applications dealing with video require detecting and tracking moving objects.
When the objects of interest have a well defined shape,
template matching or more sophisticated classifiers can be used to directly segment the objects from the image.
These techniques work well for well defined objects such as vehicles but are difficult to implement for non-rigid objects such as human bodies.
A more common approach for detecting people in a video sequence is to detect foreground pixels,
for example via Gaussian mixture models~\cite{Reddy_AVSS_2010,StaufferAndGrimson2000}.
However, current techniques typically have one major disadvantage:
shadows tend to be classified as part of the foreground.
This happens because shadows share the same movement patterns and have a similar magnitude of intensity change
as that of the foreground objects~\cite{NadimiAndBhanu2004}.

\begin{figure*}[b!]
  \centering
  \includegraphics[width=0.65\textwidth,height=0.45\textwidth]{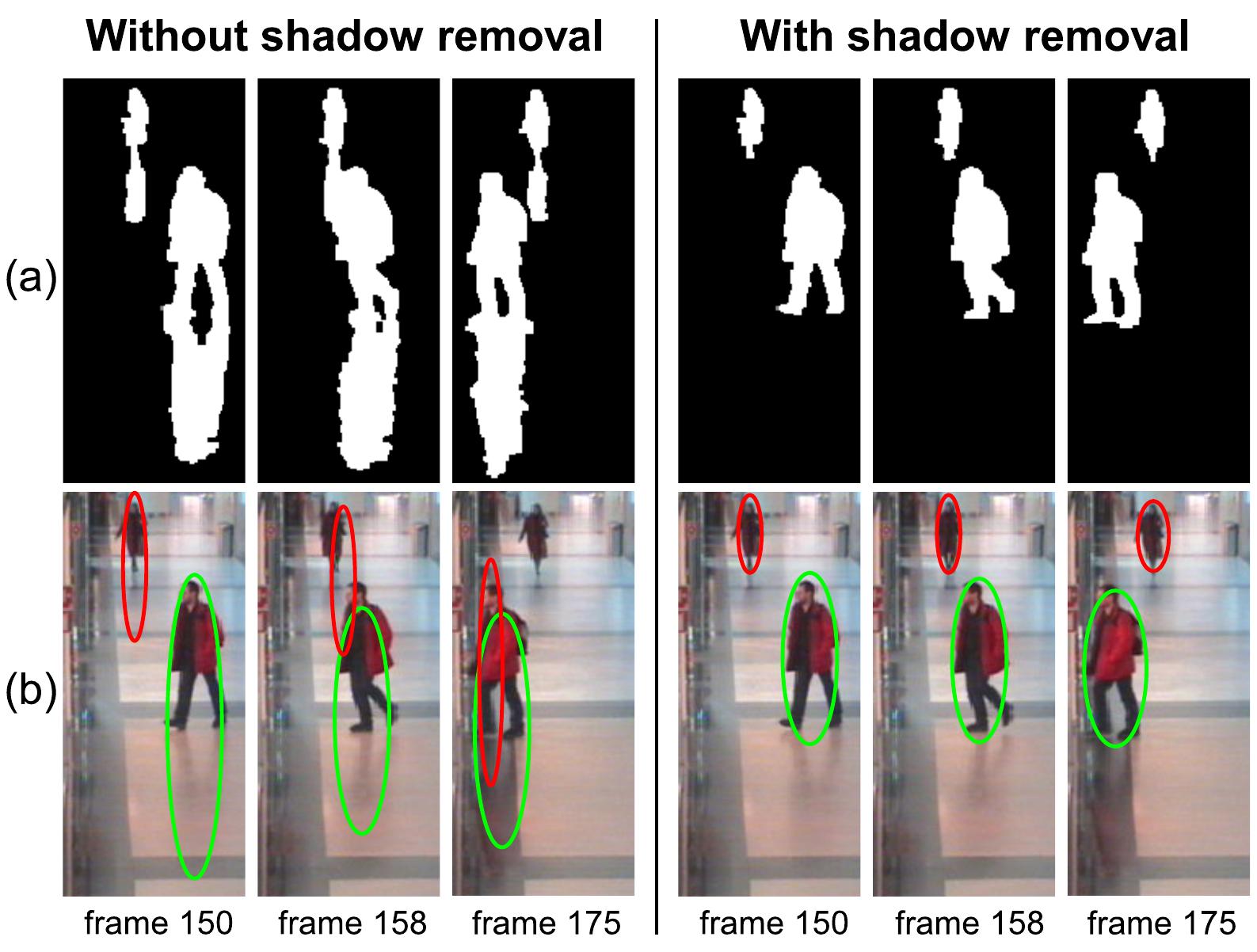}
  \caption
    {
    A case where the correct tracking trajectory can only be obtained when shadows are removed.
    {\bf (a)}~Foreground masks without shadow removal (left group) and with shadow removal (right group).
    {\bf (b)}~Tracking results, with the boundary of each object represented by an ellipse with a unique colour.
    Without shadow removal the boundaries are unreliable, resulting in the loss of tracking of one of the objects.
    }
  \label{fig:shadow_merging}
\end{figure*}

Since cast shadows can be as big as the actual objects, their incorrect classification as foreground results in inaccurate detection and decreases
tracking performance. Example scenarios where detection and tracking performance are affected include: {\bf (i)}~several people are merged together
because of their cast shadows, {\bf (ii)}~the inclusion of shadow pixels decreases the reliability of the appearance model for each person, increasing
the likelihood of tracking loss. Both scenarios are illustrated in Figure~\ref{fig:shadow_merging}. As such, removing shadows has become an
unavoidable step in the implementation of robust tracking systems~\cite{MitraEtAl2007}.

The last review of shadow detection and removal techniques was done in 2003 by Prati et al.~\cite{PratiEtAl2003}.
The review categorised shadow detection methods in an algorithm-based taxonomy.
From each class, the authors selected one algorithm to do a comparative evaluation.
The main conclusion was that only the simplest methods were suitable for generalisation,
but in almost every particular scenario the results could be significantly improved by adding assumptions.
As a consequence, there was no single robust shadow detection technique
and it was better for each particular application to develop its own technique according to the nature of the scene.

Since the review by Prati et al.~\cite{PratiEtAl2003}, many new methods have been proposed.
In this paper, we present an updated survey and an extensive comparative evaluation.
In our survey, we categorise the cast shadow detection methods published during the last decade into four feature-based categories.
From each category, we select and implement one or more prominent methods.
We compare the selected methods in a detailed evaluation, involving qualitative, quantitative and applied experiments,
to show the strengths and weaknesses of each method.

We continue the paper as follows.
In Section~\ref{sec:review}, we describe the main features that are used for detecting moving cast shadows
and use them to categorise recent shadow detection methods.
In Section~\ref{sec:methods}, a subset of the methods is chosen for implementation, with each selected method explained in detail.
In Section~\ref{sec:results} we perform an extensive quantitative and qualitative comparison of the selected methods.
In Section~\ref{sec:discussion} we present our concluding remarks.

\section{Detecting Moving Cast Shadows}
\label{sec:review}

Prati et al.~\cite{PratiEtAl2003} classified shadow detection methods in an algorithm-based taxonomy.
As a secondary classification, they mentioned the types of features used by each method among three broad classes:
spectral, spatial and temporal features.
We have observed that the choice of features has greater impact on shadow detection results compared to the choice of algorithms.
Therefore, we present a feature-based taxonomy with a secondary mention of the types of algorithms.
Furthermore, we divide spectral features into intensity, chromacity and physical properties.
We divide spatial features into geometry and textures.

In Section~\ref{sec:review_features}, we explain how each type of feature can be used to detect shadows. In Section~\ref{sec:review_taxonomy}, we
categorise recent shadow detection methods into a feature-based taxonomy.

\subsection{Useful features for shadow detection}
\label{sec:review_features}

Most of the following features are more useful for detecting shadows when the frame, which contains objects and their shadows, can be compared with an
estimation of the background, which has no objects or moving cast shadows. This review focuses in detecting shadows produced by moving objects in
video sequences, where it is reasonable to assume that a clear view of the background can be obtained, or that the background can be estimated even in
the presence of foreground objects~\cite{ToyamaEtAl1999,ReddyEtAl2011}.

\subsubsection{Intensity}

The simplest assumption that can be used to detect cast shadows is that regions under shadow become darker as they are blocked from the illumination
source. Furthermore, since there is also ambient illumination, there is a limit on how much darker they can become. These assumptions can be used to
predict the range of intensity reduction of a region under shadow, which is often used as a first stage to reject non-shadow
regions~\cite{LeoneAndDistante2007,HuangAndChen2009,HsiehEtAl2003,TianEtAl2005,ZhangEtAl2006}. However, there are no methods which rely primarily on
intensity information for discriminating between shadows and objects.

\subsubsection{Chromacity}

Most shadow detection methods based on spectral features use colour information. They use the assumption that regions under shadow become darker but
retain their chromacity. Chromacity is a measure of colour that is independent of intensity. For instance, after a green pixel is covered by shadow it
becomes dark-green, which is darker than green but has the same chromacity. This colour transition model where the intensity is reduced but the
chromacity remains the same is normally referred to as colour constancy~\cite{HorprasertEtAl1999} or linear attenuation~\cite{HuangAndChen2009}.
Methods that use this model for detecting shadows often choose a colour space with better separation between chromacity and intensity than the RGB
colour space (eg.~HSV~\cite{CucchiaraEtAl2003}, c1c2c3~\cite{SalvadorEtAl2004}, YUV~\cite{ChenEtAl2010}, normalised RGB~\cite{CavallaroEtAl2005}), or
a combination of them~\cite{SunAndLi2010}. Most of these methods are simple to implement and computationally inexpensive. However, because they make
comparisons at the pixel-level, they are susceptible to noise~\cite{PratiEtAl2003}. Furthermore, they are sensitive to strong illumination changes and
fail with strong shadows~\cite{NadimiAndBhanu2004}.

\subsubsection{Physical properties}

The linear attenuation model assumes that the illumination source produces pure white light~\cite{NadimiAndBhanu2004}, which is often not the case. In
outdoors environments, the two major illumination sources are the sun (white light) and the light reflected from the sky (blue light). Normally, the
white light from the sun dominates any other light source. When the sun's light is blocked, the effect of sky illumination increases, shifting the
chromacity of the region under shadow towards the blue component.
Nadimi and Bhanu~\cite{NadimiAndBhanu2004} proposed a dichromatic model
which takes into account both illumination sources to better predict the colour change of shadowed regions.
Further work has been done to create more general non-linear
attenuation models accounting for various illumination conditions in both indoor and outdoor
scenarios~\cite{MartelBrissonAndZaccarin2008,HuangAndChen2009}.
Alternatively, some methods address the non-linear attenuation problem by learning the
appearance that every pixel has under shadow without explicitly proposing an attenuation
model~\cite{PorikliAndThornton2005,LiuEtAl2007,MartelBrissonAndZaccarin2007,JoshiAndPapanikolopoulos2008}. These methods that try to model or learn
the specific appearance of shadow pixels are typically referred to as physical approaches. By learning or modelling particular scenarios, these
methods tend to be more accurate than chromacity methods (direct comparisons are reported in~\cite{LiuEtAl2007,JoshiAndPapanikolopoulos2008}).
However, since they are still limited to spectral properties, their main disadvantage involves dealing with objects having similar chromacity to that
of the background~\cite{HuangAndChen2009}.

\subsubsection{Geometry}

In theory, the orientation, size and even shape of the shadows can be predicted with proper knowledge of the illumination source, object shape and the
ground plane. Some methods use this information to split shadows from
objects~\cite{HsiehEtAl2003,YoneyamaEtAl2003,NicolasAndPinel2006,FangEtAl2008,ChenAndAggarwal2010}. The main advantage of geometry features is that
they work directly in the input frame; therefore, they do not rely on an accurate estimation of the background reference. However, methods that use
geometry features impose scene limitations such as: specific object types, typically pedestrians (ie.~standing
people)~\cite{HsiehEtAl2003,ChenAndAggarwal2010} or vehicles~\cite{YoneyamaEtAl2003,FangEtAl2008}; requiring objects and shadows to have different
orientation~\cite{HsiehEtAl2003,ChenAndAggarwal2010}; and assuming a unique light source~\cite{NicolasAndPinel2006} or a flat background
surface~\cite{FangEtAl2008}. Additionally, current geometry-based methods are not designed to deal with objects having multiple shadows or (except
for~\cite{HsiehEtAl2003}) multiple objects detected as a single foreground blob.

\subsubsection{Textures}

Some methods exploit the fact that regions under shadow retain most of their texture. Texture-based shadow detection methods typically follow two
steps: {\bf (1)}~selection of candidate shadow pixels or regions, and {\bf (2)}~classification of the candidate pixels or regions as either foreground
or shadow based on texture correlation. Selection of the shadow candidates is done with a weak shadow detector, usually based on spectral features.
Then, each shadow candidate is classified as either object or shadow by correlating the texture in the frame with the texture in the background
reference. If a candidate's texture is similar in both the frame and the background, it is classified as shadow. Various methods perform this
correlation with various techniques (eg.~normalised cross-correlation~\cite{TianEtAl2005}, gradient or edge
correlation~\cite{JavedAndShah2002,XuEtAl2005,SaninEtAl2010}, orthogonal transforms~\cite{ZhangEtAl2006}, Markov or conditional random
fields~\cite{WangEtAl2006,QinEtAl2010}, Gabor filtering~\cite{LeoneAndDistante2007}). Texture correlation is a potentially powerful method for
detecting shadows as textures are highly distinctive, do not depend on colours, and are robust to illumination changes. However, texture-based shadow
detection methods tend to be slow as they often have to compute one or several neighbourhood comparisons for each pixel.

\subsubsection{Temporal features}

Finally, since moving cast shadows share the same movement pattern as the objects that produce them, the same temporal consistency filters that have
been applied to the objects can be applied to the shadows~\cite{CavallaroEtAl2005,NadimiAndBhanu2004,LiuEtAl2007,NicolasAndPinel2006}. This filtering
usually enhances the detection results by keeping only the pixels that are consistent in time. However, as with the intensity features, there are no
methods which rely primarily on temporal features for shadow detection.

\subsection{Taxonomy of recent shadow detection methods}
\label{sec:review_taxonomy}

We categorised the shadow detection methods published during the last decade according to their feature choice. Although some of the methods use more
than one feature, we take into account the feature that makes the dominant contribution to the detection results or the novelty of the paper. As
mentioned in Section~\ref{sec:review_features}, intensity features are used mainly as a first step for detecting shadows, and temporal features are
mainly used for filtering the detection results. Therefore, all the reviewed methods were classified into one of four categories:
\begin{inparaenum}[\bf (i)]
  \item chromacity-based methods,
  \item physical methods,
  \item geometry-based methods,
  \item texture-based methods.
\end{inparaenum}
Our taxonomy is detailed in Table~\ref{tab:review}. The highlighted methods were chosen for the comparative evaluation and are explained in
Section~\ref{sec:methods}.

\begin{table*}[tb!]
  \renewcommand{\baselinestretch}{1.06}\small\normalsize
  \centering
  \footnotesize
  \begin{tabularx}{1\textwidth}{|l|l|l|l|X|}
    \hline
    \multicolumn{5}{|c|}{\bf Chromacity-based methods}\\
    \hline
    \textit{Paper}               &\textit{Colour space} &\textit{Level} &\textit{Spatial verification} &\textit{Temporal verification}\\
    \hline
    {\bf Cucchiara et al.~2003~\cite{CucchiaraEtAl2003}}
                                 &HSV                   &Pixel          &---                           &---\\
    Salvador et al.~2004~\cite{SalvadorEtAl2004}   
                                 &c1c2c3                &Window         &No internal shadows           &---\\
    Cavallaro et al.~2005~\cite{CavallaroEtAl2005}  
                                 &Normalised RGB        &Pixel          &No internal shadows           &Tracking life-span\\
    Chen et al.~2010~\cite{ChenEtAl2010}       
                                 &YUV                   &Pixel          &Morphology                    &---\\
    Sun and Li 2010~\cite{SunAndLi2010}       
                                 &HSI and c1c2c3        &Pixel          &Morphology                    &---\\
    \hline
  \end{tabularx}
  \\
  ~
  \\
  ~
  \\
  \begin{tabularx}{1\textwidth}{|l|l|l|X|}
    \hline
    \multicolumn{4}{|c|}{\bf Physical methods}\\
    \hline
    \textit{Paper}                         &\textit{Model} &\textit{Learning}      &\textit{Spatial or temporal cues}\\
    \hline
    Nadimi and Bhanu 2004~\cite{NadimiAndBhanu2004}           
                                           &Dichromatic    &---                    &Spatio-temporal test\\
    Porikli and Thornton 2005~\cite{PorikliAndThornton2005}       
                                           &---            &Shadow flow            &---\\
    Liu et al.~2007~\cite{LiuEtAl2007}                  
                                           &---            &Gaussian mixture model &Markov random fields and tracking\\
    Martel-Brisson and Zaccarin~2007~\cite{MartelBrissonAndZaccarin2007} 
                                           &---            &Gaussian mixture model &---\\
    Martel-Brisson and Zaccarin 2008~\cite{MartelBrissonAndZaccarin2008} 
                                           &General        &Kernel based           &Gradients (direction)\\
    Joshi and Papanikolopoulos 2008~\cite{JoshiAndPapanikolopoulos2008} 
                                           &---            &Semisupervised (SVM)   &Edges\\
    {\bf Huang and Chen 2009~\cite{HuangAndChen2009}}
                                           &General        &Gaussian mixture model &Gradients (attenuation)\\
    \hline
  \end{tabularx}
  \\
  ~
  \\
  ~
  \\
  \begin{tabularx}{1\textwidth}{|l|l|l|l|X|}
    \hline
    \multicolumn{5}{|c|}{\bf Geometry-based methods}\\
    \hline
    \textit{Paper}                &\textit{Objects} &\textit{Blob segmentation} &\textit{Main cue}     &\textit{Other cues}\\
    \hline
    {\bf Hsieh et al.~2003~\cite{HsiehEtAl2003}}
                                  &People           &Via head detection         &Orientation           &Intensity and location\\
    Yoneyama et al.~2003~\cite{YoneyamaEtAl2003}
                                  &Vehicles         &---                        &2D models             &Vanishing point\\
    Nicolas and Pinel~2006~\cite{NicolasAndPinel2006} 
                                  &Any              &---                        &Light source          &Temporal filter\\
    Fang et al.~2008~\cite{FangEtAl2008}        
                                  &Vehicles         &---                        &Wave transform        &Spectral\\
    Chen and Aggarwal 2010~\cite{ChenAndAggarwal2010} 
                                  &People           &---                        &Log-polar coordinates &Colour and oriented gradients\\
    \hline
  \end{tabularx}
  \\
  ~
  \\
  ~
  \\
  \begin{tabularx}{1\textwidth}{|l|l|l|X|}
    \hline
    \multicolumn{4}{|c|}{\bf Texture-based methods}\\
    \hline
    \textit{Paper}                     &\textit{Weak detector}       &\textit{Texture correlation}    &\textit{Correlation level}\\
    \hline
    Javed and Shah 2002~\cite{JavedAndShah2002}         
                                       &Colour segmentation          &Gradient direction correlation  &Medium region\\
    Xu et al.~2005~\cite{XuEtAl2005}               
                                       &---                          &Static edge correlation         &Pixel\\
    Tian et al.~2005~\cite{TianEtAl2005}             
                                       &Intensity range              &Normalised cross-correlation    &Pixel\\ 
    Wang et al.~2006~\cite{WangEtAl2006}             
                                       &---                          &Intensity and edge DCRF filter  &Small region\\
    Zhang et al.~2006~\cite{ZhangEtAl2006}            
                                       &---                          &Orthogonal transforms           &Small region\\
    {\bf Leone and Distante 2007~\cite{LeoneAndDistante2007}}
                                       &Photometric gain             &Gabor filter                    &Small region\\
    Zhang et al.~2007~\cite{ZhangEtAl2007}            
                                       &Intensity constraint         &Ratio edge test                 &Small region\\
    Nghiem et al.~2008~\cite{NghiemEtAl2008}
                                       &Chromacity based             &Intensity reduction ratio       &Small region\\
    Shoaib et al.~2009~\cite{ShoaibEtAl2009}
                                       &---                          &Gradient background subtraction &Pixel\\
    Pei and Wang 2009~\cite{PeiAndWang2009}
                                       &---                          &PCA based                       &Small region\\
    {\bf Sanin et al.~2010~\cite{SaninEtAl2010}}
                                       &Chromacity based             &Gradient direction correlation  &Large region\\
    Nakagami and Nishitani 2010~\cite{NakagamiAndNishitani2010} 
                                       &---                          &Walsh transform domain          &Small region\\
    Panicker and Wilscy 2010~\cite{PanickerAndWilscy2010}    
                                       &---                          &Foreground edge detection       &Pixel\\
    Qin et al.~2010~\cite{QinEtAl2010}              
                                       &Shadow colour model          &Local ternary pattern MRF       &Small region\\
    \hline
  \end{tabularx}
  \caption
    {
    Taxonomy of recently published shadow detection methods. Methods are sorted by year in the first column and grouped by category. The additional
    columns show secondary classifications within each category. Chromacity-based methods are subdivided according their colour space, level of
    granularity and additional spatial or temporal verification. Physics-based methods are subdivided according to their physical shadow model,
    learning algorithm and additional spatial or temporal cues. Geometry-based methods are subdivided according to their supported object type,
    whether they support multiple objects per blob, their main geometrical cue and additional cues. Texture-based methods are subdivided according to
    their weak shadow detector, texture correlation method and the size of the regions used in the correlation. The highlighted methods were chosen
    for the comparative evaluation.
    }
  \label{tab:review}
  \renewcommand{\baselinestretch}{1.2}\small\normalsize
\end{table*}

\clearpage

\section{Methods Selected for Implementation}
\label{sec:methods}

For practical reasons, only a subset of the reviewed methods was implemented.
In this section, we detail a selection of one or more prominent methods from each category.
These methods were implemented and used for the comparative evaluation in Section~\ref{sec:results}.

\subsection{Chromacity-based method}

Among the chromacity methods, the most important factor is to choose a colour space with a separation of intensity and chromacity. Several colour
spaces such as HSV~\cite{CucchiaraEtAl2003}, c1c2c3~\cite{SalvadorEtAl2004} and normalised RGB~\cite{CavallaroEtAl2005} have proved to be robust for
shadow detection~\cite{ShanEtAl2007}.
We chose the HSV approach proposed by Cucchiara et al.~\cite{CucchiaraEtAl2003},
since that colour space provides a natural separation between chromacity and luminosity.
This shadow detection method has been widely used in surveillance applications
(eg.~\cite{MaddalenaAndPetrosino2008,ForczmanskiAndSeweryn2010}).
Since the value (V) is a direct measure of intensity,
pixels in the shadow should have a lower value than pixels in the background.
Following the chromacity cues, a shadow cast on background does not change its hue (H)
and the authors noted that shadows often lower the saturation (S) of the points.
Therefore, a pixel $p$ is considered to be part of a shadow if the following three conditions are satisfied:

\begin{align}
  \beta_{1} \leq \left({F_{p}^{V}} / {B_{p}^{V}}\right) &\leq \beta_{2}\\
  \left(F_{p}^{S} - B_{p}^{S}\right) &\leq \tau_{S}\\
  \left|F_{p}^{H} - B_{p}^{H}\right| &\leq \tau_{H}
\end{align}

\noindent
where {\small $F_{p}^{C}$} and {\small $B_{p}^{C}$} represent the component values, {\small $C$}, of HSV for the pixel position $p$ in the frame ($F$)
and in the background reference image ($B$), respectively. $\beta_{1}$, $\beta_{2}$, $\tau_{S}$ and $\tau_{H}$ represent thresholds that are optimised
empirically. Working with alternative colour spaces may produce different but not necessarily better results~\cite{ShanEtAl2007}. However, extending
the pixel-level based analysis to an observation window improves results by countering pixel-level noise~\cite{SalvadorEtAl2004}. In our
implementation of the HSV method we used a 5-by-5 observation window rather than treating each pixel separately.

\subsection{Physical method}

Research in physical models for cast shadow removal has been done incrementally.
The more recent papers are extensions of previous physical models,
typically removing some assumptions and improving on previous results.
We chose a recent approach by Huang and Chen~\cite{HuangAndChen2009}
which does not make prior assumptions about the light sources and ambient illumination,
and reports better results than similar methods. For a pixel~$p$, given the vector
from shadow to background value denoted as $v(p)$, the colour change is modelled using the 3D colour feature $x(p) = [\alpha(p), \theta(p), \phi(p)]$.
Here, $\alpha(p)$ represents the illumination attenuation, while $\theta(p)$ and $\phi(p)$ indicate the direction of $v(p)$ in spherical coordinates:

\begin{align}
  \alpha(p) &= ||v(p)|| / ||BG(p)||\\
  \theta(p) &= \arctan \left( v^G(p)/v^R(p) \right)\\
  \phi(p) &= \arccos \left( v^B(p)/||v(p)|| \right)
\end{align}

\noindent
where $BG(p)$ is the background value at the pixel $p$, and the superscripts $R$, $G$, $B$ indicate the components in the RGB colour space. This
colour feature describes the appearance variation induced by the blocked light sources on shaded regions. The model is learned in an unsupervised way.
First, a weak shadow detector identifies pixels in the foreground that have reduced luminance and different saturation from that of the background.
Then, the attenuation of these candidate shadow pixels is used to update a Gaussian mixture model of the 3D colour features, penalising the learning
rate of pixels with larger gradient intensities than the background, which are more likely to be foreground objects. Finally, posterior probabilities
of the model are used to classify each pixel in the foreground as object or shadow.

\subsection{Geometry-based method}

Most geometry methods assume that each foreground blob contains a single object and shadow,
which is not guaranteed in many computer vision applications.
For this reason, we chose the method proposed by Hsieh et al.~\cite{HsiehEtAl2003},
which separates the blobs into individual objects before doing the geometric analysis.
As in most geometry-based methods, their work assumes that the objects of interest are persons and that their shadows have a
different orientation. First, they analyse the vertical peaks on each blob to detect potential heads, and then use this information to split the blobs
into person-shadow pairs. Given a person-shadow region $R$, its centre of gravity $(\bar{x}, \bar{y})$ and orientation $\theta$ are found as follows:

\begin{align}
  (\bar{x}, \bar{y}) &= \left(\frac{1}{|R|}\sum_{(x,y) \epsilon R} x, \frac{1}{|R|}\sum_{(x,y) \epsilon R} y\right)\\
  \theta &= \frac{1}{2}\arctan\left(\frac{2\mu_{1,1}}{\mu_{2,0} - \mu_{0,2}}\right)
\end{align}

\noindent
where $|R|$ is the area of the region in pixels, and $\mu_{p,q}$ are the correspondent central moments. The point below the centre of gravity with the
maximum vertical change is considered to be the pixel where the shadow begins, and a segment oriented according to $\theta$ is used to roughly split a
candidate shadow region $R_2$. Then, the following Gaussian model is built from the pixels in $R_2$:

\begin{small}
\begin{equation}
  G(s,t,g) = \exp\left[-\left(\frac{w_s s^2}{\sigma_s^2} + 
                              \frac{w_t t^2}{\sigma_t^2} + 
                              \frac{w_g (g - \mu_g)^2}{\sigma_g^2}\right)\right]
\end{equation}
\end{small}

\noindent
where $s$ and $t$ are the elliptical coordinates of each pixel and $g$ its intensity (ie.~$g = I(s,t)$), and $w$ and $\sigma^2$ are the weight and
variance of each component in the Gaussian. This model summarises the intensity of the shadow pixels and includes the coordinates where the shadow is
expected to be. Once the model is built, every pixel in the original region $R$ is classified as object or shadow, according to whether it agrees with
the Gaussian model or not.

\subsection{Small region (SR) texture-based method}

Texture-based methods present the greatest diversity among the various categories.
As a representative of methods which use small regions (or neighbourhoods) to correlate textures,
we chose the method proposed by Leone and Distante~\cite{LeoneAndDistante2007},
as it correlates textures using Gabor functions.
Region-level correlation is more robust than pixel-level correlation and Gabor functions can provide optimal joint localisation in the
spatial/frequency domains~\cite{RandenAndHusoy1999}. As in the majority of the texture-based shadow detection methods, the method first creates a mask
with the potential shadow pixels in the foreground. Then, if the textures of small region centered at each pixel are correlated to the background
reference, the pixels are classified as shadow. In this case, the potential shadow pixels are found using a photometric gain measure which gives
higher probability to pixels whose intensity is lower than the background. The texture analysis is performed by projecting a neighbourhood of pixels
onto a set of Gabor functions with various bandwidths, orientations and phases, and the matching between frame and background is found using Euclidean
distance. Since a full Gabor dictionary may be expensive to compute, a sub-dictionary with the most representative kernels can be first found using
the matching pursuit strategy~\cite{BergeaudAndMallat1995}.

\subsection{Large region (LR) texture-based method}

The problem of using small regions is that they are not guaranteed to contain significant textures.
Sanin et al.~\cite{SaninEtAl2010} proposed using colour features to first create large candidate shadow regions
(ideally containing whole shadow areas),
which are then discriminated from objects using gradient-based texture correlation.

The purpose of the first step is to create regions of pixels as large as possible which contain shadow pixels or object pixels, but not both.
Candidate shadow pixels are found using intensity and chromacity features in the HSV colour space as in~\cite{CucchiaraEtAl2003},
but adapting the thresholds to ensure high detection accuracy.
Classifying all shadow pixels as shadows is important in this step because it will impose the upper bound of the final detection accuracy.
Misclassifying object pixels as shadows is not an issue since they are later discriminated using texture features.
The connected components of shadow pixels are used as candidate shadow regions.
Some candidate shadow regions may contain both shadow and object pixels.
To avoid this scenario, we implemented the optional step described in~\cite{SaninEtAl2010}
where regions are split using edges that occur in the foreground but not in the background reference.

In the second step, the texture for each candidate region is correlated between the frame and the background reference. Since shadows tend to preserve
the underlying textures, shadow regions should have a high texture correlation. First, for each candidate region, the gradient magnitude $\left|
\bigtriangledown_{p} \right|$ and gradient direction $\theta_{p}$ at each pixel $p = (x,y)$ are calculated using:

\begin{align}
  \left|\bigtriangledown_{p}\right| &= \sqrt{{\bigtriangledown_{x}}^2 + {\bigtriangledown_{y}}^2}\\
  \theta_{p} &= \operatorname{arctan2} \left({\bigtriangledown_{y}} / {\bigtriangledown_{x}}\right)
\end{align}

\noindent
where $\bigtriangledown_{y}$ is the vertical gradient (difference in intensity between the pixel and the pixel in the next row), while
$\bigtriangledown_{x}$ is the horizontal gradient. The function $\operatorname{arctan2}(\cdot)$ is a variant of $\arctan(\cdot)$ that returns an angle
in the full angular range $[-\pi,\pi]$, allowing the gradient direction to be treated as a true circular variable. Only the pixels with $\left|
\bigtriangledown_{p} \right|$ greater than a certain threshold $\tau_{m}$ are taken into account to avoid the effects of noise, which is stronger in
the smooth regions of the frame.

Since the gradient direction is a circular variable, the difference has to be calculated as an angular distance. For each pixel $p = (x,y)$ that was
selected due to significant magnitude, the difference in gradient direction between the frame $F$ and the background reference $B$ is calculated
using:

\begin{equation}
  \Delta \theta_{p} = \arccos \frac{\bigtriangledown_{x}^F \bigtriangledown_{x}^B +
                                    \bigtriangledown_{y}^F \bigtriangledown_{y}^B}
                                   {\sqrt{({\bigtriangledown_{x}^F}^2 + {\bigtriangledown_{y}^F}^2)
                                          ({\bigtriangledown_{x}^B}^2 + {\bigtriangledown_{y}^B}^2)}}
\end{equation}

The gradient direction correlation between the frame and the background is estimated using:

\begin{equation}
  c = \frac{\sum_{p = 1}^{n} \operatorname{H} (\tau_{a} - \Delta \theta_{p})}{n}
\end{equation}

\noindent
where $n$ is the number of pixels selected in the candidate shadow region and $H(\cdot)$ is the unit step function which, in this case, evaluates to
$1$ if the angular difference is less than or equal to the threshold $\tau_{a}$, and $0$ otherwise. In essence, $c$ is the fraction of pixels in the
region whose gradient direction is similar in both the frame and the background. If $c$ is greater than threshold $\tau_{c}$, the candidate region is
considered a shadow region and it is removed from the foreground mask.
\section{Comparative Evaluation}
\label{sec:results}

In this section, we present two sets of experiments to compare the performance of the five methods%
\footnote{C++ source code for the implemented methods can be obtained from~\url{http://arma.sourceforge.net/shadows/}}
selected in Section~\ref{sec:review}.
The first experiment is a direct measure of shadow detection performance.
We show both quantitative and qualitative results.
The second experiment is an applied empirical study that shows the improvement in tracking performance after using each of the five compared methods.
Below, we describe the sequences used for the experiments and the steps for creating the ground truth frames.
In the following subsections, we explain each experiment in detail and present the results.

The sequences used in our experiments are summarised in Table~\ref{tab:datasets}. We used a wide range of scenes with variations in the type and size
of objects and shadows. The first six sequences were introduced in~\cite{PratiEtAl2003}\footnote{\url{http://cvrr.ucsd.edu/aton/shadow/}}
and~\cite{MartelBrissonAndZaccarin2008}\footnote{\url{http://vision.gel.ulaval.ca/~CastShadows/}}, and have been widely used for testing shadow
detection performance. The last entry summarises 25 sequences from the CAVIAR
dataset\footnote{\url{http://homepages.inf.ed.ac.uk/rbf/CAVIARDATA1/}}.
Each sequence presents a different challenge for the shadow detection methods
to tests their robustness. The \textit{Campus} sequence is a particularly noisy outdoor sequence where some of the shadows are extremely long. The
\textit{Hallway} sequence has a textured background and the size of the people changes significantly according to their distance to the camera. The
\textit{Highway 1} and \textit{Highway 3} sequences show a traffic environment with two different lighting conditions and vehicle sizes. In
particular, the \textit{Highway 3} has very small vehicles which could be misclassified as shadows. The \textit{Lab} and \textit{Room} indoor
sequences show two laboratory rooms in two different perspectives and lighting conditions.
From the CAVIAR dataset, we selected 25 sequences that show people walking in various patterns inside a shopping mall.

\begin{table*}[tb!]

  \centering
  \scriptsize
  
  \begin{tabular}{llccccccc}
    \toprule
    \multicolumn{2}{c}{}
    &\begin{minipage}{45pt}
      \vspace{2pt} \centerline{\bf Campus} \vspace{1pt}
      \includegraphics[width=\columnwidth,height=0.75\columnwidth]{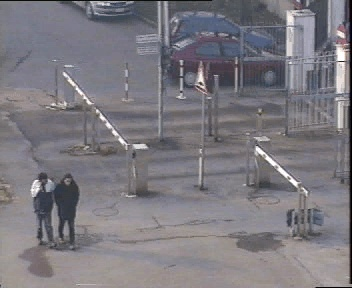}
    \end{minipage}
    &\begin{minipage}{45pt}
      \vspace{2pt} \centerline{\bf Hallway} \vspace{1pt}
      \includegraphics[width=\columnwidth,height=0.75\columnwidth]{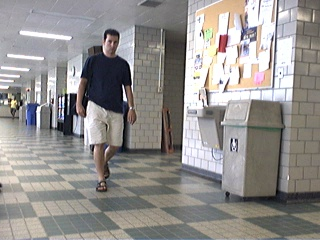}
    \end{minipage}
    &\begin{minipage}{45pt}
      \vspace{2pt} \centerline{\bf Highway 1} \vspace{1pt}
      \includegraphics[width=\columnwidth,height=0.75\columnwidth]{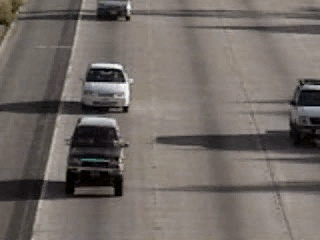}
    \end{minipage}
    &\begin{minipage}{45pt}
      \vspace{2pt} \centerline{\bf Highway 3} \vspace{1pt}
      \includegraphics[width=\columnwidth,height=0.75\columnwidth]{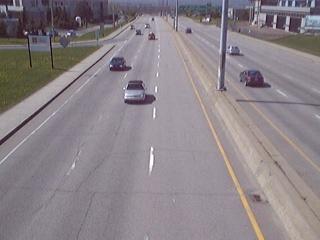}
    \end{minipage}
    &\begin{minipage}{45pt}
      \vspace{2pt} \centerline{\bf Lab} \vspace{1pt}
      \includegraphics[width=\columnwidth,height=0.75\columnwidth]{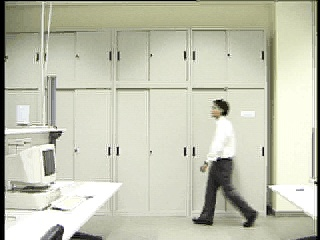}
    \end{minipage}
    &\begin{minipage}{45pt}
      \vspace{2pt} \centerline{\bf Room} \vspace{1pt}
      \includegraphics[width=\columnwidth,height=0.75\columnwidth]{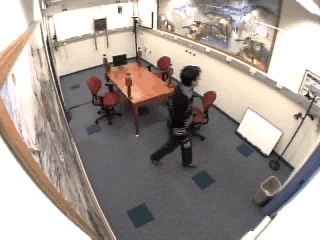}
    \end{minipage}
    &\begin{minipage}{45pt}
      \vspace{2pt} \centerline{\bf CAVIAR} \vspace{1pt}
      \includegraphics[width=\columnwidth,height=0.75\columnwidth]{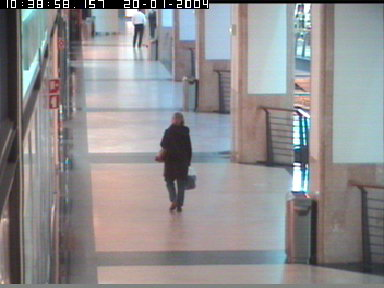}
    \end{minipage}\\
    \midrule[\heavyrulewidth]
    \multirow{3}{*}{\rotatebox[origin=c]{90}{\it Frames}}
    &{\it Number}    &1179            &1800           &440            &2227             &887            &300            &1388 ($\times$25)\\
    &{\it Labelled}  &53              &13             &8              &7                &14             &22             &45 ($\times$25)\\
    &{\it Size}      &352$\times$288  &320$\times$240 &320$\times$240 &320$\times$240   &320$\times$240 &320$\times$240 &384$\times$288\\
    \midrule
    \multirow{3}{*}{\rotatebox[origin=c]{90}{\it Scene}}
    &{\it Type}      &outdoor         &indoor         &outdoor        &outdoor          &indoor         &indoor         &indoor\\
    &{\it Surface}   &asphalt         &\bf{textured}  &asphalt        &asphalt          &\bf{white}     &carpet         &\bf{reflective}\\
    &{\it Noise}     &\bf{high}       &medium         &medium         &medium           &low            &medium         &medium\\
    \midrule
    \multirow{3}{*}{\rotatebox[origin=c]{90}{\it Objects}}
    &{\it Type}      &vehicles/people &people         &vehicles       &vehicles         &people/other   &people         &people\\
    &{\it Size}      &medium          &\bf{variable}  &large          &\bf{small}       &medium         &medium         &medium\\
    &{\it Speed}     &5-10            &5-15           &\bf{30-35}     &20-25            &10-15          &2-5            &1-4\\
    \midrule
    \multirow{3}{*}{\rotatebox[origin=c]{90}{\it Shadows}}
    &{\it Size}      &\bf{very large} &medium         &large          &small            &medium         &\bf{large}     &medium\\
    &{\it Strength}  &weak            &weak           &\bf{strong}    &\bf{very strong} &weak           &medium         &medium\\
    &{\it Direction} &horizontal      &multiple       &horizontal     &horizontal       &\bf{multiple}  &\bf{multiple}  &\bf{vertical}\\
    \bottomrule
  \end{tabular}
  \caption
    {
    Image sequences used in the comparative evaluation.
    Sequences are described in terms of:
    number and size of frames and number of examples with manually labelled shadows;
    scene type, surface and noise;
    type, size (relative to the frame size) and speed (in pixels/frame) of the objects;
    and size, strength and direction of the shadows.
    The main features or challenges presented by each sequences are highlighted using bold text.
    Note that the last column summarises the properties of 25 sequences from the CAVIAR dataset,
    all of which are different recordings of the same scene.
    }
  \label{tab:datasets}
  ~
\end{table*}

\begin{figure*}[tb!]
  \centering
  \begin{minipage}{0.8\textwidth}
  \begin{minipage}{0.235\textwidth}
    \includegraphics[width=\textwidth]{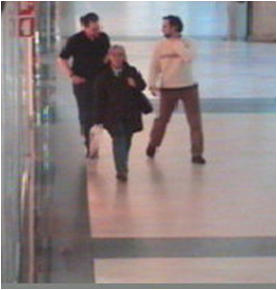}
    \centerline{\footnotesize\bf (a)}
  \end{minipage}
  \hfill
  \begin{minipage}{0.235\textwidth}
    \includegraphics[width=\textwidth]{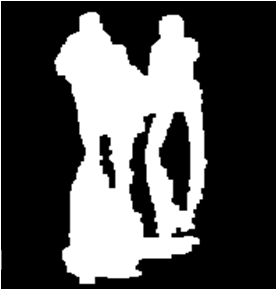}
    \centerline{\footnotesize\bf (b)}
  \end{minipage}
  \hfill
  \begin{minipage}{0.235\textwidth}
    \includegraphics[width=\textwidth]{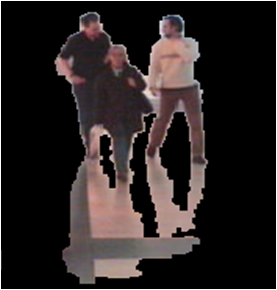}
    \centerline{\footnotesize\bf (c)}
  \end{minipage}
  \hfill
  \begin{minipage}{0.235\textwidth}
    \includegraphics[width=\textwidth]{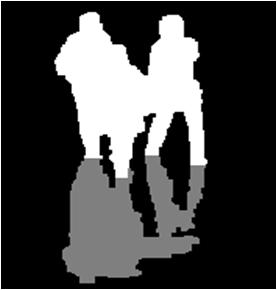}
    \centerline{\footnotesize\bf (d)}
  \end{minipage}
  \end{minipage}
  \caption
    {
    Ground truth creation steps:
    {\bf (a)} original frame;
    {\bf (b)} foreground mask obtained with GMM;
    {\bf (c)} frame superimposed on foreground mask;
    {\bf (c)} labelled mask with object pixels in white and shadow pixels in grey.
    }
  \label{fig:ground_truth}
\end{figure*}

Manually labelling shadow pixels in a frame is much more difficult than labelling object pixels.
For instance, it may be easy to determine which pixels belong to the people in the frame in Figure~\ref{fig:ground_truth}~(a),
but it is difficult to tell which pixels correspond to their cast shadows.
For this reason, we first used the standard Gaussian mixture model (GMM) foreground extraction method from OpenCV 2.0~\cite{Bradski2008}
to extract a foreground mask on each frame.
The resulting mask should contain only object and shadow pixels.
By superimposing the original frame on the mask, labelling pixels in the mask as object or cast shadow becomes straightforward.
The labelling process is summarised in Figure~\ref{fig:ground_truth}.
The labelled masks, along with the original frames and the backgrounds estimated with the GMM method
are a further contribution of this paper\footnote{The ground truth masks can be obtained from~\url{http://arma.sourceforge.net/shadows/}}.
The original sequences can be used for learning-based methods which require all the intermediate frames.

\newpage

\subsection{Shadow detection performance}
First, we measured the shadow detection performance of each method in every test sequence.
We then gradually decreased the colour information of each sequence to test the dependency of each method on colour features.
Finally, we show qualitative results for each sequence and summarise their observed behaviour.
All the methods were faithfully implemented as described in Section~\ref{sec:review}.
The thresholds for all methods were selected to obtain the best overall performance on the test sequences,
using one setting for all sequences (ie.~no specific tuning for each sequence).

\subsubsection{Quantitative results}
To test the shadow detection performance of the five methods we used the two metrics proposed by Prati et al.~\cite{PratiEtAl2003},
namely shadow detection rate ($\eta$) and shadow discrimination rate ($\xi$):

\begin{align}
  \eta &= \frac{{TP}_{S}}{{TP}_{S} + {FN}_{S}}\\
  \xi  &= \frac{{TP}_{F}}{{TP}_{F} + {FN}_{F}}
\end{align}

\noindent
where ${TP}$ and ${FN}$ stand for true positive and false negative pixels with respect to either shadows ($S$) or foreground objects ($F$).
The shadow detection rate is concerned with labelling the maximum number of cast shadow pixels as shadows.
The shadow discrimination rate is concerned with maintaining the pixels that belong to the moving object as foreground.
In this paper we often use the average of the two rates as a single performance measure.

\begin{figure*}[tb!]
  \centering
  \includegraphics[width=\textwidth]{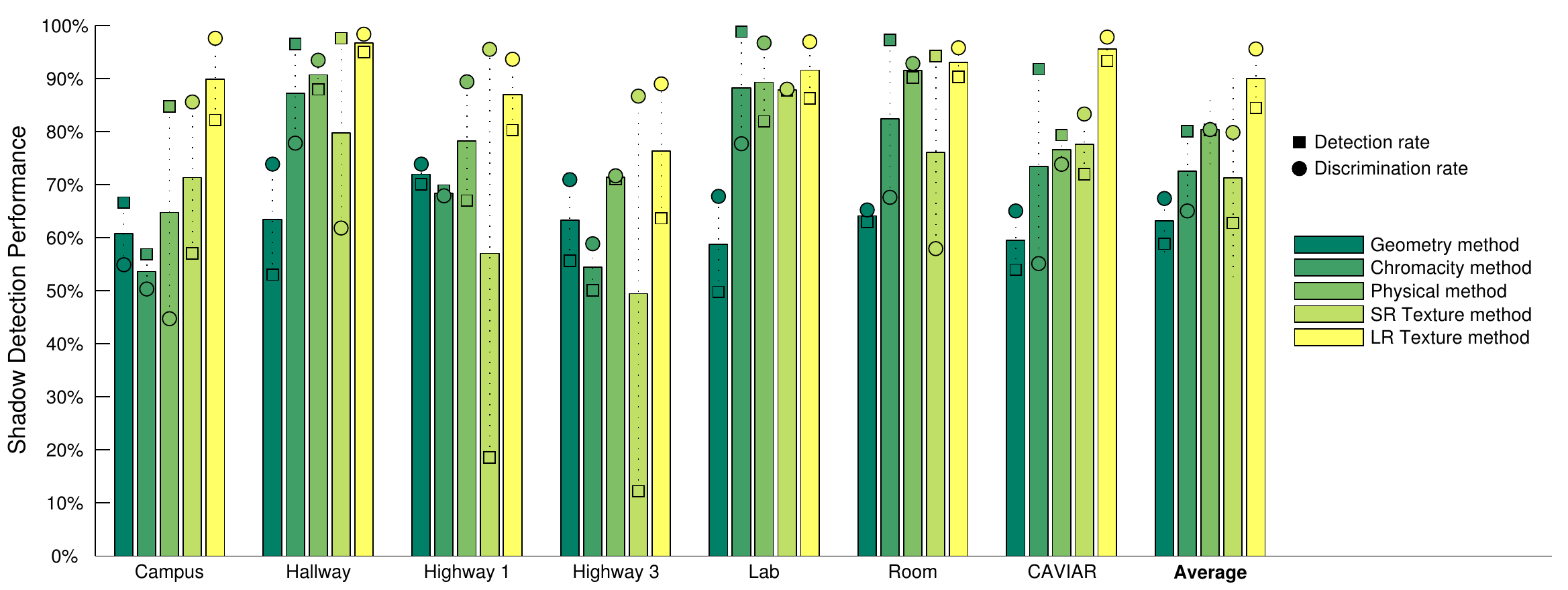}
  \caption
    {
    Comparison of shadow detection results by sequence.
    The performance measure is the average between the shadow detection and discrimination rates.
    Each bar represents the average performance with a vertical segment whose length indicates the average difference between the detection rate
    (marked with a square) and the discrimination rate (marked with a circle).
    }
  \label{fig:accuracy_results}
\end{figure*}

Figure~\ref{fig:accuracy_results} shows the average shadow detection and discrimination rates on each test sequence.
Each bar represents the average of the detection and discrimination rates on each sequence,
while the individual detection and discrimination rates are indicated by a square and a circle, respectively.
In all cases, the large region texture-based method performs considerably better than all the others,
obtaining high values for both the detection and discrimination rates in all sequences.
We discuss these sequence-related results more deeply in the qualitative results section.

The average processing time per frame of each shadow detection method is shown in Table~\ref{tab:speed}.
Times are shown in milliseconds, obtained on a 32-bit Intel CPU running at 2.6~GHz.
Four of the methods, excluding the small region texture-based,
have the same asymptotic complexity of constant operations per pixel.
The chromacity-based method was the quickest to implement and run.
The geometry and physical-based methods need more operations for calculating the central moments and updating the shadow models, respectively.
The large region texture-based method requires extra steps to generate the candidate shadow regions and calculate the gradients for each pixel,
and is the slowest of these four.
The amount of operations per pixel in the small region texture-based method depends on the size
and number of kernels used for correlating the textures.
This small region texture-based method has considerably higher computational load than the rest,
even when the number of kernels and their size are minimised.

\begin{table*}[!tb]
  \centering
  \small
  \begin{tabular}{lccccc}
    \toprule
                      &{\bf Chromacity} &{\bf Geometry} &{\bf Physical} &{\bf SR Textures}    &{\bf LR Textures}\\
    \midrule
    \textit{Campus}   &8.72            &9.44          &10.00         &156.46 (48.81)      &20.76\\
    \textit{Hallway}  &11.28           &8.91          &12.81         &223.64 (77.37)      &21.59\\
    \textit{Highway1} &10.73           &24.75         &16.93         &341.32 (116.28)     &34.71\\
    \textit{Highway3} &6.82            &6.49          &7.15          &120.36 (37.07)      &11.75\\
    \textit{Lab}      &8.95            &17.68         &15.34         &253.82 (82.48)      &22.73\\
    \textit{Room}     &7.14            &8.41          &8.51          &144.87 (47.07)      &16.25\\
    \textit{Caviar}   &10.82           &13.94         &14.08         &243.07 (82.98)      &24.70\\
    {\bf Average}     &\bf{9.21}       &\bf{12.80}    &\bf{12.12}    &\bf{211.93 (70.30)} &\bf{21.78}\\
    \bottomrule
  \end{tabular}
  \caption
    {
    Average frame processing time (in milliseconds) per sequence for various shadow detection methods.
    The first value shown for the small region (SR) texture-based method corresponds to the processing time using 48 Gabor kernels.
    This configuration was used in the previous experiments for maximum accuracy.
    The second value in brackets represents the time when using a reduced set of 16 Gabor kernels (for faster processing).
    }
  \label{tab:speed}
  ~
\end{table*}

The large region texture-based method was designed firstly to increase the effectiveness of texture-based features by selecting large regions,
and secondly to use spectral features to improve results in the absence of significant textures.
However, it is hard to conclude how well the method achieves these two goals by simply observing the shadow detection performance on various sequences.
For this reason, we tested the performance of each method while gradually decreasing the colour information of each sequence
until reaching greyscale frames.
This experiment is used for two purposes:
{\bf (1)} to make a fair comparison between both texture-based methods when colour information is lost
and thus determine which one is using texture information more effectively;
{\bf (2)} to observe how the colour information is used when available, and how dependent are the methods on colour features.

\begin{figure*}[!tb]
  \centerline{\includegraphics[width=0.6\textwidth]{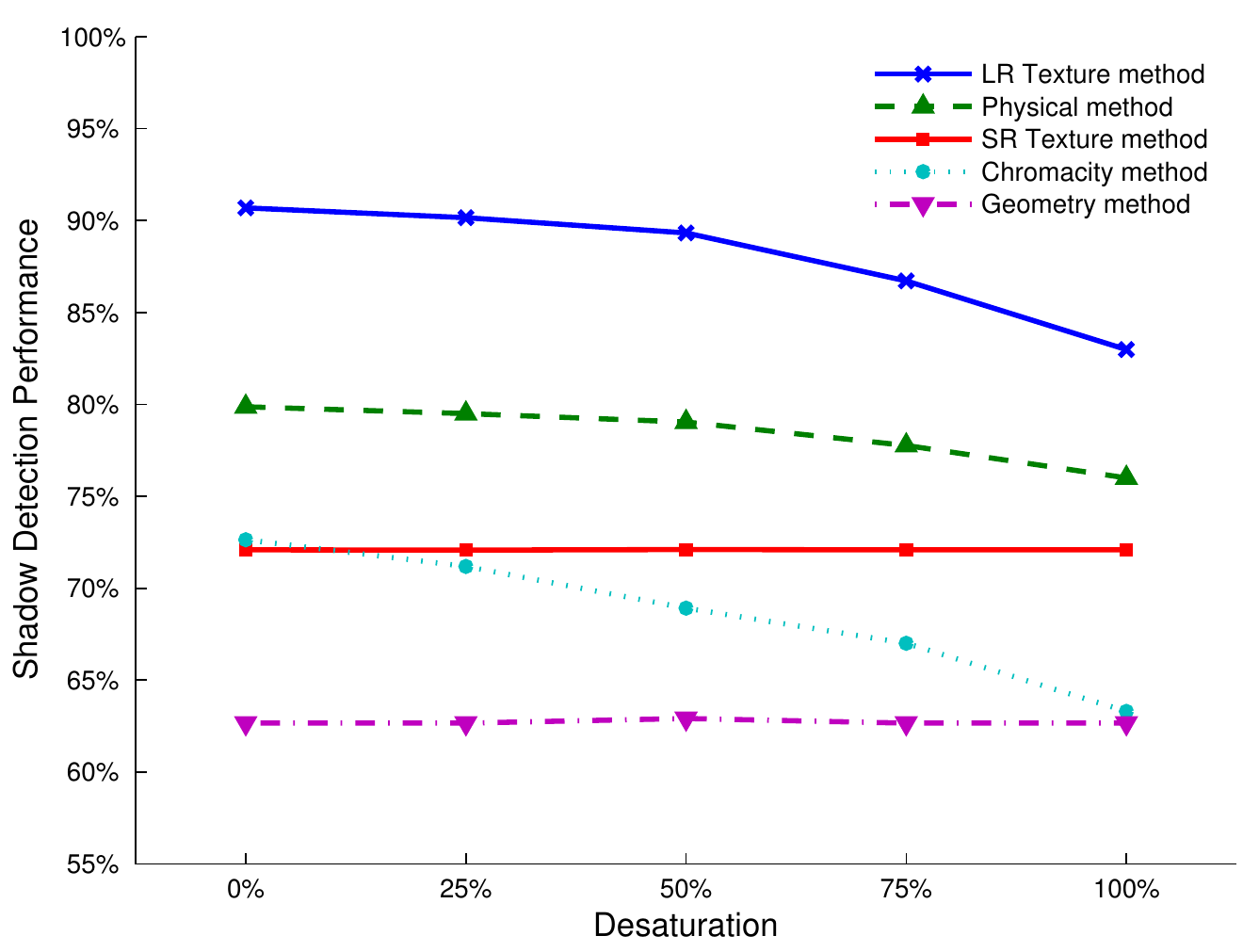}}
  \caption
    {
    Shadow detection performance when the colour information is gradually reduced. Each line shows the average performance of a shadow detection
    method ranging from the original sequences (0\% desaturated) to the greyscale-converted sequences (100\%~desaturated).
    }
  \label{fig:desaturation_results}
\end{figure*}

Figure~\ref{fig:desaturation_results} shows the average performance of each method across all sequences as the colour information is manually
decreased. The performance results indicate the average between shadow detection and discrimination rates. The desaturation rate indicates the
reduction in colour information used to modify the frames as follows:

\begin{equation}
  F_\mathit{desat} = (1 - \lambda) F_\mathit{orig} + \lambda F_\mathit{grey}
\end{equation}

\noindent
where $\lambda$ is the desaturation rate, $F_\mathit{orig}$ is the original frame and $F_\mathit{grey}$ is the frame converted to greyscale.
In other words, when the desaturation is 0\% the original frames are used,
when it is 100\% the frames are converted to greyscale,
and the rest are interpolations between the two, created by gradual blending to avoid colour distortions.
Note that alternative ways of reducing the colour information can also be used.
For example, it is possible to represent the images in HSV colour space and gradually decrease the saturation (S) component to remove colour.

Several things can be concluded from this figure. First, as expected, methods which use colour features perform better when all the colour information
is available, and their performance drops as the frames are desaturated. Second, the decrease in performance of the physical method is lower since it
is locally adaptive; the performance of the large region texture-based method also decreases, but with a magnitude lower than that of the chromacity
method, since it is globally adaptive. Third, without colours the large region texture-based method depends on textures while performing considerably
better than the small region texture-based method. Last, the large region texture-based method performs better than all others regardless of the
degree of desaturation; while it uses colour information to improve performance, without such information its performance is nevertheless still good.

\subsubsection{Qualitative results}

\begin{figure*}[tb!]

  \centering
  \footnotesize

  \begin{tabular}{@{}m{1em}qq@{\hspace{2pt}}!{\vrule width 1pt}@{\hspace{2pt}}qqqqq}
    \rotatebox[origin=c]{90}{\it Campus}
    &\includegraphics[width=0.1345\textwidth]{qual_campus.png}
    &\includegraphics[width=0.1345\textwidth]{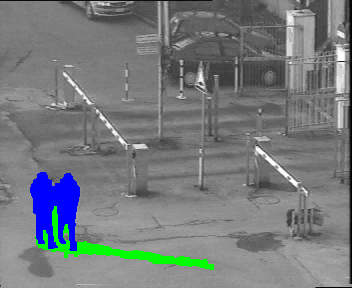}
    &\includegraphics[width=0.1345\textwidth]{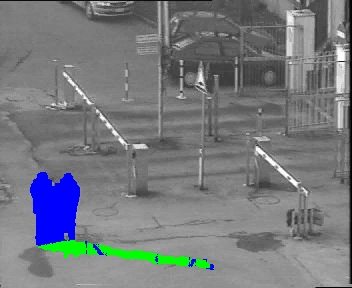}
    &\includegraphics[width=0.1345\textwidth]{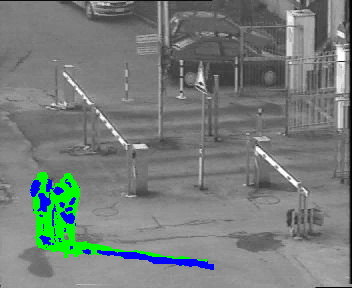}
    &\includegraphics[width=0.1345\textwidth]{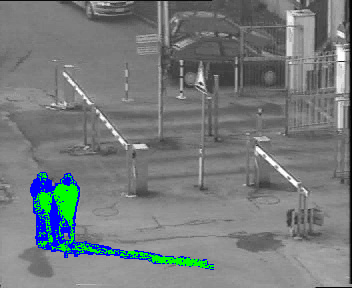}
    &\includegraphics[width=0.1345\textwidth]{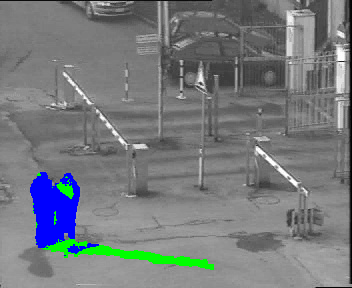}
    &\includegraphics[width=0.1345\textwidth]{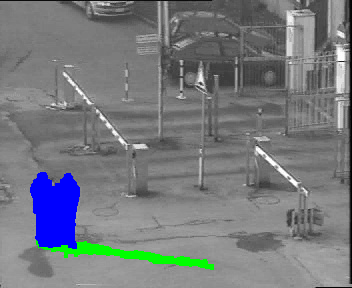}
    \\
    \rotatebox[origin=c]{90}{\it Hallway}
    &\includegraphics[width=0.1345\textwidth]{qual_hallway.png}
    &\includegraphics[width=0.1345\textwidth]{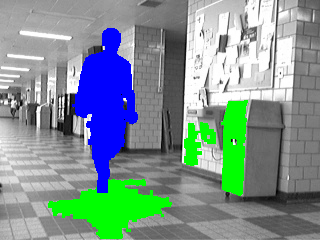}
    &\includegraphics[width=0.1345\textwidth]{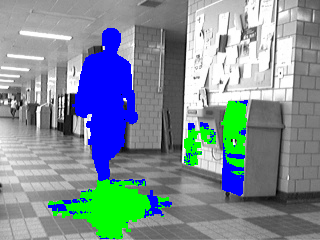}
    &\includegraphics[width=0.1345\textwidth]{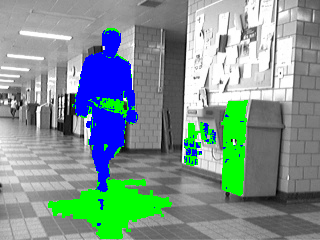}
    &\includegraphics[width=0.1345\textwidth]{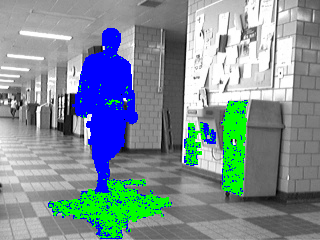}
    &\includegraphics[width=0.1345\textwidth]{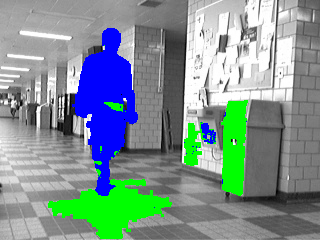}
    &\includegraphics[width=0.1345\textwidth]{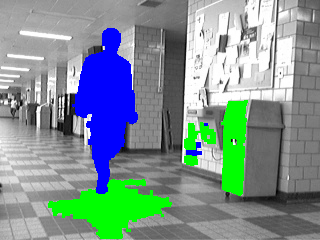}
    \\
    \rotatebox[origin=c]{90}{\it Highway 1}
    &\includegraphics[width=0.1345\textwidth]{qual_highway1.png}
    &\includegraphics[width=0.1345\textwidth]{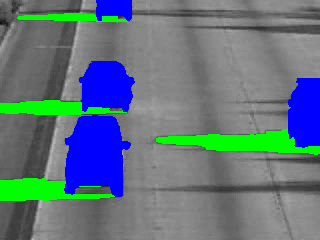}
    &\includegraphics[width=0.1345\textwidth]{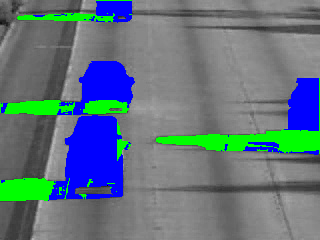}
    &\includegraphics[width=0.1345\textwidth]{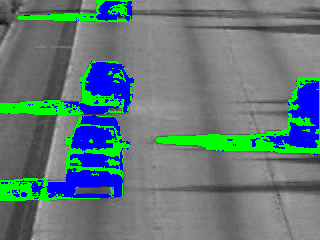}
    &\includegraphics[width=0.1345\textwidth]{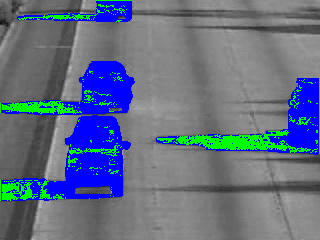}
    &\includegraphics[width=0.1345\textwidth]{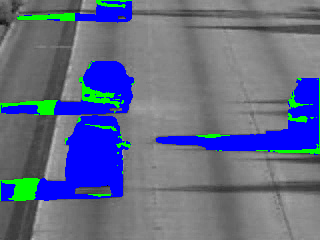}
    &\includegraphics[width=0.1345\textwidth]{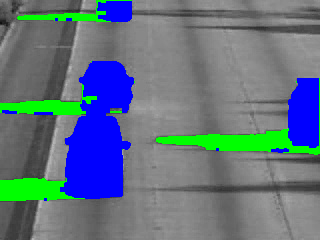}
    \\
    \rotatebox[origin=c]{90}{\it Highway 3}
    &\includegraphics[width=0.1345\textwidth]{qual_highway3.png}
    &\includegraphics[width=0.1345\textwidth]{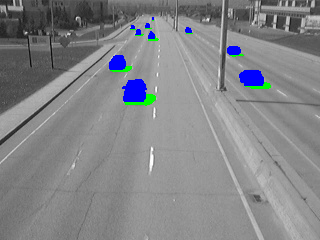}
    &\includegraphics[width=0.1345\textwidth]{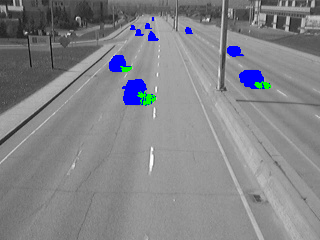}
    &\includegraphics[width=0.1345\textwidth]{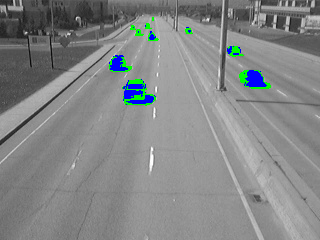}
    &\includegraphics[width=0.1345\textwidth]{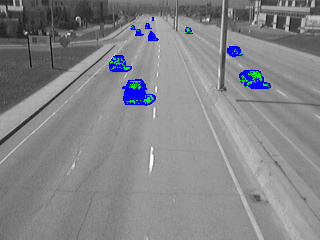}
    &\includegraphics[width=0.1345\textwidth]{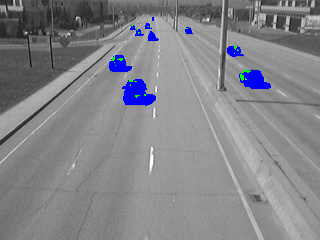}
    &\includegraphics[width=0.1345\textwidth]{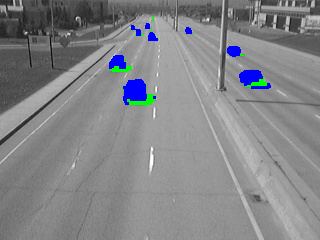}
    \\
    \rotatebox[origin=c]{90}{\it Lab}
    &\includegraphics[width=0.1345\textwidth]{qual_lab.png}
    &\includegraphics[width=0.1345\textwidth]{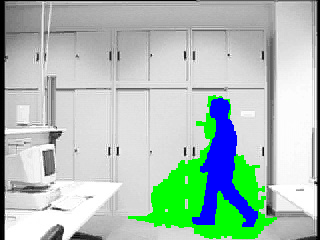}
    &\includegraphics[width=0.1345\textwidth]{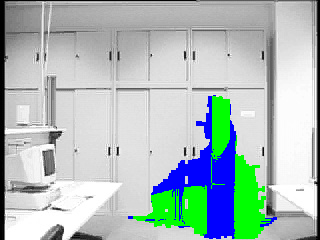}
    &\includegraphics[width=0.1345\textwidth]{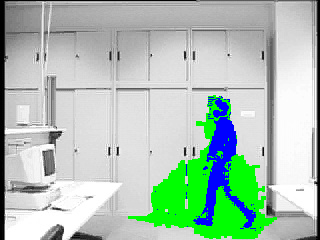}
    &\includegraphics[width=0.1345\textwidth]{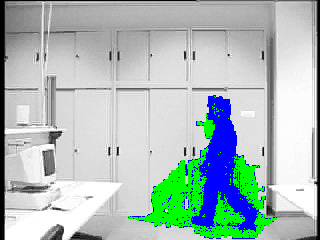}
    &\includegraphics[width=0.1345\textwidth]{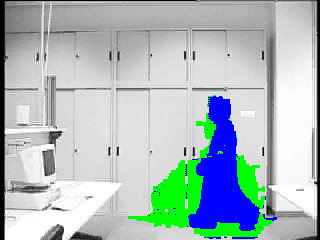}
    &\includegraphics[width=0.1345\textwidth]{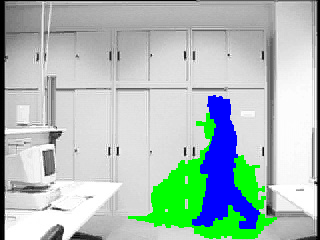}
    \\
    \rotatebox[origin=c]{90}{\it Room}
    &\includegraphics[width=0.1345\textwidth]{qual_room.png}
    &\includegraphics[width=0.1345\textwidth]{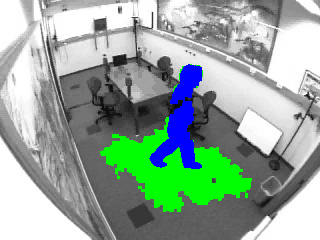}
    &\includegraphics[width=0.1345\textwidth]{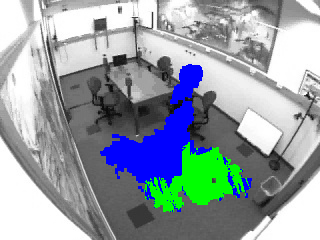}
    &\includegraphics[width=0.1345\textwidth]{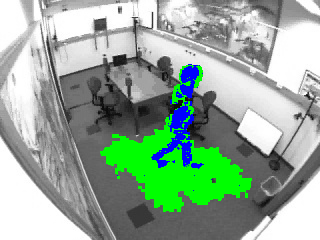}
    &\includegraphics[width=0.1345\textwidth]{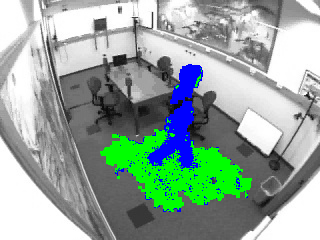}
    &\includegraphics[width=0.1345\textwidth]{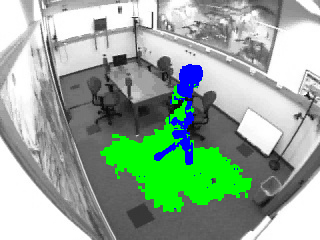}
    &\includegraphics[width=0.1345\textwidth]{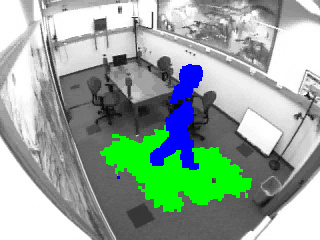}
    \\
    \rotatebox[origin=c]{90}{\it CAVIAR}
    &\includegraphics[width=0.1345\textwidth]{qual_caviar.png}
    &\includegraphics[width=0.1345\textwidth]{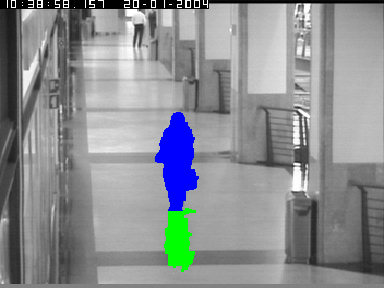}
    &\includegraphics[width=0.1345\textwidth]{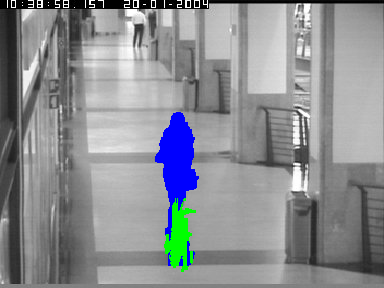}
    &\includegraphics[width=0.1345\textwidth]{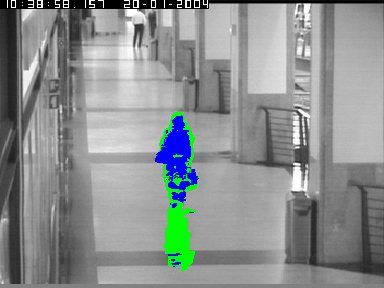}
    &\includegraphics[width=0.1345\textwidth]{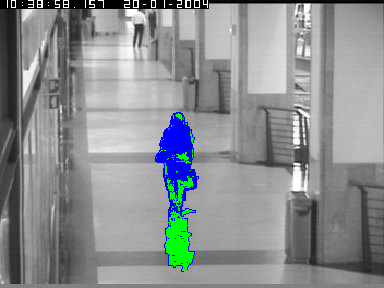}
    &\includegraphics[width=0.1345\textwidth]{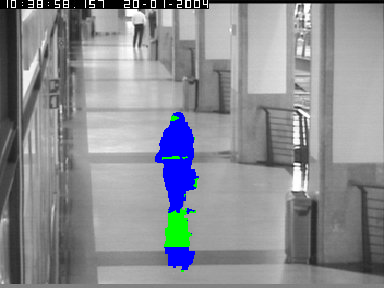}
    &\includegraphics[width=0.1345\textwidth]{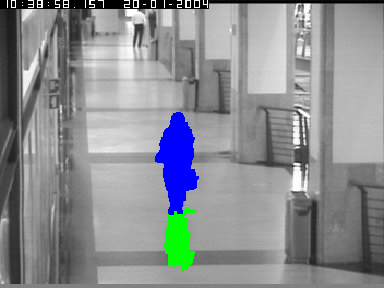}
    \\
    &{\bf Original frame}
    &{\bf Ground truth}
    &{\bf Geometry}
    &{\bf Chromacity}
    &{\bf Physical}
    &{\bf SR Textures}
    &{\bf LR Textures}
    \\
  \end{tabular}
  \caption
    {
    Qualitative shadow detection results.
    The first column shows an example frame for each sequence.
    The second column show the expected results where object pixels are marked in blue and shadow pixels are marked in green.
    The remaining columns show the observed results for each method.
    The geometry-based method works well for long shadows with a well defined orientation (\textit{Campus} example)
    but does not work well for spread shadows (\textit{Lab} example).
    Both the chromacity-based and physical methods are affected by pixel-level noise,
    and fail when objects are darker and have similar colours to the background (\textit{Campus} example),
    although the physical method adapts better to the different scenes.
    The small region (SR) texture-based method fails for pixels located in non-textured neighbourhoods (\textit{Highway 1} example).
    The large region (LR) texture-based method works reasonably well in all cases.
    }
  \label{fig:qualitative_results}
\end{figure*}

We show the qualitative results in Figure~\ref{fig:qualitative_results}. The first column shows an example frame for each sequence, the second column
shows the expected results, where foreground pixels are marked in blue and shadow pixels are marked in green. The remaining columns show the observed
results with each shadow detection method. For symmetry, we show results for all sequences and all methods. However, it is important to note that the
geometry-based method was designed for pedestrian shadow detection and, apart from mentioning the limitation, we do not take into account the
sequences with objects other than pedestrians when comparing to this particular method. In general, as was also shown in the qualitative results, the
large region texture-based method performs better in all the examples.

Specific observations can be also done for each method. The geometry-based method only works when each shadow has a unique orientation which differs
to the object's orientation, as happens in the \textit{Campus} example, but fails when the shadows have the same orientations as the objects or when
shadows have multiple directions as in the \textit{Lab} example. The chromacity-based method is affected by pixel-level noise, and fails when objects
are darker and have similar colours to the background as in the \textit{Campus} example. Although the physical method also uses chromacity features,
it has the ability to adapt to each scene and it is less likely to fail when the chromacity-based method fails. The small region texture-based method
works well in scenes with textured backgrounds but fails for pixels located in non-textured neighbourhoods (as can be seen in the \textit{Highway 1}
example). Finally, the large region texture-based method works well in most cases, although it can distort the object contours as in the \textit{Lab}
example.

The results from both the quantitative experiments and qualitative observations are summarised in Table~\ref{tab:qualitative_results}.
We assign scores to each method according to several criteria.
The geometry-based method has strong assumptions regarding the object and shadow shape,
but when these assumptions are met the method works well independently on the quality of spectral and texture features.
The chromacity-based method is simple and fast,
and as Prati et al.~\cite{PratiEtAl2003} concluded,
its few assumptions lead it to work reasonably well in most scenes.
However, it has a strong trade-off between shadow detection and discrimination.
The physical method reduces the limitations of the chromacity-based method,
provided there are sufficient examples to learn and adapt to the appearance of moving cast shadow in the background.
However, as the chromacity-based method, it is sensitive to pixel-level noise and scenes with low saturated colours.
The small region texture-based method is robust to various illumination conditions and is easy to implement,
but it requires the background to be textured and needs a costly texture correlation operation per pixel.
The large region texture-based method is not sensitive to pixel-level noise,
and it is independent to the type of shadows, objects and scenes.
It presents the best results at the cost of additional computational load.

\newpage

\begin{table*}[!tb]

  \centering
  \footnotesize
  
  \begin{tabular}{lccccc}
    \toprule
                                                &{\bf Geometry} &{\bf Chromacity} &{\bf Physical} &{\bf SR Textures} &{\bf LR Textures}\\
    \midrule[\heavyrulewidth]
    {\bf 1. Scene independence}                 &medium         &medium           &medium         &medium            &high\\
    \midrule
    {\bf 2. Object independence}                &low            &high             &high           &high              &high\\
    \midrule
    {\bf 3. Shadow independence}                &low            &high             &high           &high              &high\\
    \midrule
    {\bf 4. Penumbra detection}                 &medium         &low              &medium         &high              &high\\
    \midrule
    {\bf 5. Robustness to noise}                &medium         &low              &medium         &high              &high\\
    \midrule
    {\bf 6. Detection/discrimination trade-off} &low            &high             &medium         &high              &low\\
    \midrule
    {\bf 7. Computational load}                 &low            &low              &low            &high              &medium\\
    \bottomrule
  \end{tabular}
  \caption{
    Summary of the comparative evaluation. The shadow detection methods are rated as {\it low}, {\it medium} or {\it high} according to seven
    criteria. For the first five criteria, {\it high} indicates that the method performed well in all cases. A method with low scene, object or shadow
    independence, only works well for a narrow set of scene, object or shadow types. Low penumbra detection means that a method is not appropriate for
    detecting shadow borders. Low robustness to noise means that shadow detection performance is significantly affected by scene noise. For the last
    two criteria (ie. 6 and 7) lower means better. A high detection/discrimination trade-off means that tuning a method to increase one of the rates
    will significantly decrease the other. In terms of computational load, the methods are ranked by their average processing time per frame.
  }
  \label{tab:qualitative_results}
\end{table*}

\clearpage

\subsection{Effect on tracking performance}

Given that several methods perform relatively well (ie.~detection and discrimination rates above 75\%), will pursuing higher rates translate into
improved object detection and tracking? Although measuring shadow detection accuracy pixel-by-pixel is an objective measure, applications rarely have
the need of explicitly detecting shadow pixels. Rather, shadow detection methods are typically used in tracking applications to clean the detection
results and improve tracking performance. It is important to prove that increasing the detection rate and/or discrimination rate of shadow pixels will
result in better tracking results. For this reason, we performed a second set of experiments to measure tracking performance after applying each
method to remove the shadows in the foreground masks (ie.~setting the shadow pixels to zero).

Using various tracking algorithms is important since they may use the shadow-removed foreground masks differently.
For instance, the masks can be used prior to tracking (eg.~for initialisation of particle filters~\cite{ZhouEtAl2004})
or during tracking (eg.~for updating appearance models in blob matching~\cite{FuentesAndVelastin2006,ZhouAndAggarwal2006}).
We used five tracking algorithms implemented in the video surveillance module of OpenCV v2.0~\cite{Bradski2008}:
blob matching (CC),
mean-shift tracking (MS),
mean-shift tracking with foreground feedback (MSFG),
particle filtering (PF),
and
blob matching/particle filter hybrid (CCPF).
The foreground masks are used while tracking in CC, prior to tracking in MS and PF,
and both while and prior to tracking in MSFG and CCPF.

For the experiments, we used the tracking ground truth data available for the 50 sequences in the second set of the CAVIAR dataset.
We performed 30 tracking evaluations by combining six shadow removal options
(no shadow removal, geometry-based method, chromacity-based method, physical method, small and large region texture-based methods)
with the five tracking algorithms.
The tracking performance was measured with the two metrics proposed by Bernardin and Stiefelhagen~\cite{BernardinAndStiefelhagen2008},
namely multiple object tracking accuracy (MOTA) and multiple object tracking precision (MOTP):

\begin{align}
  \text{MOTA} &= 1 - \frac{\sum_{t}(m_{t} + fp_{t} + mme_{t})}{\sum_{t}g_{t}}\\
  \text{MOTP} &= \frac{\sum_{i,t}(d_{t}^{i})}{\sum_{t}c_{t}}
\end{align}

Briefly, MOTA is an inverse measure of the number of missed objects ($m_{t}$), false positives ($fp_{t}$) and mismatches ($mme_{t}$).
The higher the MOTA, the better.
MOTP measures the average pixel distance ($d_{t}$) between the ground-truth locations of objects
and their locations according to a tracking algorithm.
The lower the MOTP, the better.
Ground truth objects and hypotheses are matched using the Hungarian algorithm \cite{Munkres1957}.

\begin{figure*}[tb!]
  \centering
  \begin{minipage}{0.49\textwidth}
    \includegraphics[width=\columnwidth]{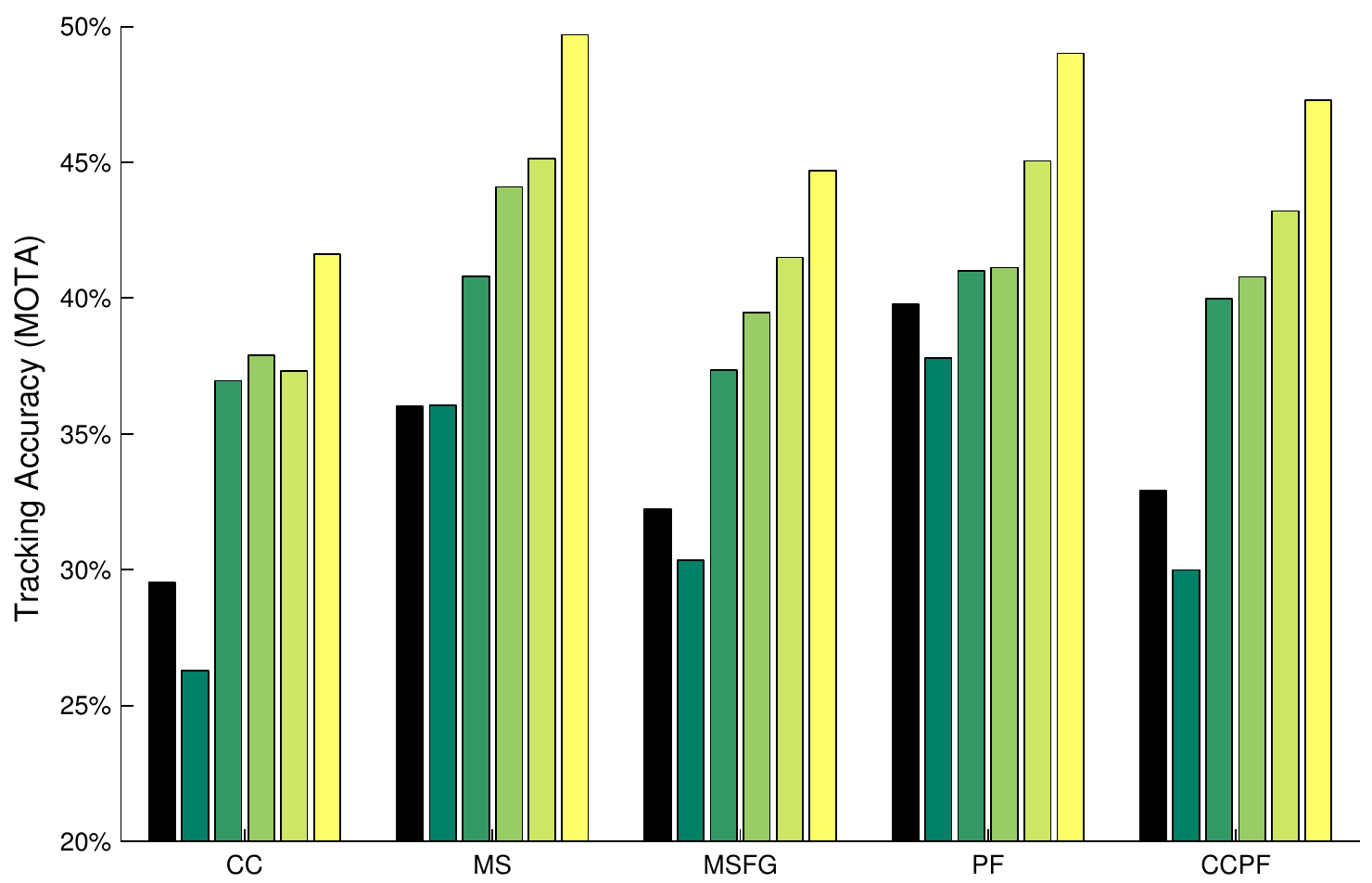}
     \centerline{\footnotesize\bf (a)}
  \end{minipage}
  \hfill
  \begin{minipage}{0.49\textwidth}
    \includegraphics[width=\columnwidth]{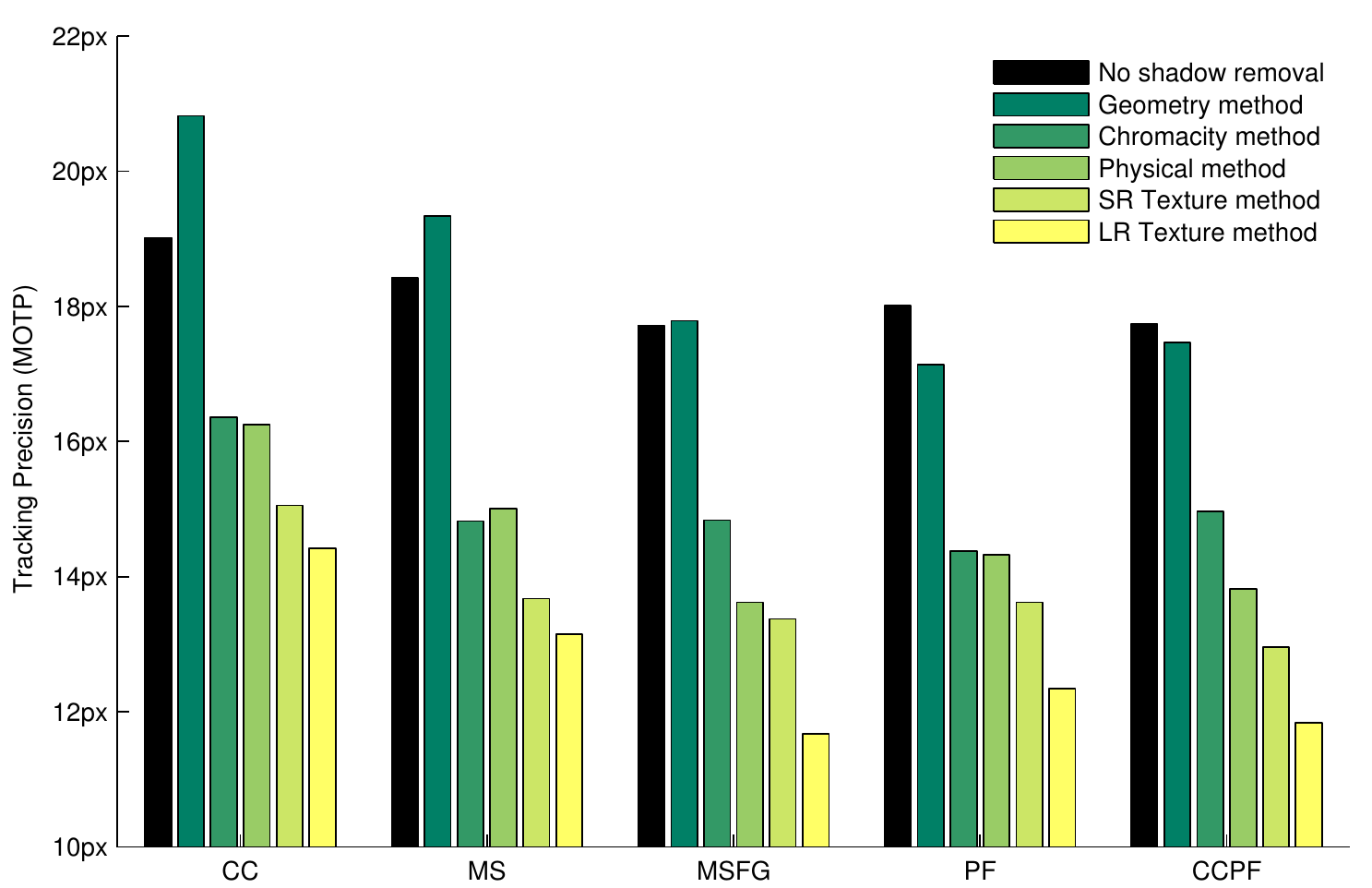}
     \centerline{\footnotesize\bf (b)}
  \end{minipage}
  \caption{\footnotesize
    Effect of shadow removal methods on:
    {\bf (a)} multiple object tracking accuracy (MOTA), where taller bars indicate better accuracy; and
    {\bf (b)} multiple object tracking precision (MOTP), where shorter bars indicate better precision.
    Results are grouped by tracking algorithm: blob matching (CC), two mean shift trackers (MS and MSFG), particle filter (PF) and hybrid tracking
    (CCPF).
  }
  \label{fig:tracking_results}
\end{figure*}

The tracking results are presented in Figure~\ref{fig:tracking_results}. Each bar represents the performance result of a particular tracking algorithm
after removing the shadows detected by one of the shadow detection methods, averaged for the 50 test sequences. In all cases, the large region
texture-based method results in better tracking performance, with considerable improvements over the next best method (small region texture-based).

Three other things can be observed in these results.
First, in most cases, tracking performance is significantly improved with shadow removal regardless of the tracking algorithm.
In some cases, the geometry-based method led to a decrease in tracking performance.
Recall that the geometry method relies on the assumption that objects and shadows have different orientations,
which is not met in the CAVIAR dataset.
Second, the improvement in tracking performance by using a better shadow remover
(eg.~large region texture-based method instead of the chromacity-based method) is comparable
to the improvement by using a better tracking algorithm (eg.~particle filter instead of blob matching).
Last, improving shadow detection performance
(as shown for the CAVIAR dataset in Figure~\ref{fig:accuracy_results})
leads to a proportional improvement in tracking performance.
Therefore, regardless of the tracking algorithm,
it is worth pursuing better shadow detection methods to obtain more accurate tracking results.

\section{Main Findings and Future Directions}
\label{sec:discussion}

In this paper we presented a review and a comparative evaluation of shadow detection methods published during the last decade.
The survey follows the work of Prati et al.~\cite{PratiEtAl2003} in 2003 but with recent publications,
a more comprehensive set of test sequences and more detailed experiments.

In the review the shadow detection methods are classified in a feature-based taxonomy.
Methods that use mainly spectral features are classified as either chromacity-based or physical methods.
Methods that use mainly spatial features are classified as either geometry-based or texture-based methods.
For the comparative evaluation, we selected a prominent method from each group,
except for the texture-based category, where we selected two:
one that uses small regions and one that uses large regions for the texture correlation.
We compared the shadow detection performance of these methods using a wide range of test sequences.
Additionally, we observed the effect of low saturation on shadow detection performance and evaluated the
practical link between shadow detection and tracking performance.

The quantitative and qualitative comparative results can serve as a guide (for both practitioners and researchers)
to select the best method for a specific purpose ---
all shadow detection approaches make different contributions and all have individual strengths and weaknesses.
Out of the selected methods, the geometry-based technique has strict assumptions and is not generalisable to various environments,
but it is a straightforward choice when the objects of interest are easy to model and their shadows have different orientation.
The chromacity-based method is the fastest to implement and run,
but it is sensitive to noise and less effective in low saturated scenes.
The physical method improves upon the accuracy of the chromacity method by adapting to local shadow models,
but fails when the spectral properties of the objects are similar to that of the background.
The small-region texture-based method is especially robust for pixels whose neighbourhood is textured,
but may take longer to implement and is the most computationally expensive.
The large-region texture-based method produces the most accurate results,
but has a significant computational load due to its multiple processing steps.

Traditionally, simple and fast shadow detection and removal methods have been favoured in computer vision applications
such as tracking systems~\cite{ForczmanskiAndSeweryn2010,LeiAndXu2006,MaddalenaAndPetrosino2008,WangAndSuter2007}.
It is hence pertinent to note that when the shadow detection performance is relatively poor
(eg.~$< 60\%$ for the geometry-based technique on the CAVIAR dataset),
shadow removal can in fact lead to tracking performance which is worse than not using shadow removal.
In contrast, more elaborate shadow detection algorithms lead to considerably better tracking performance,
regardless of the tracking algorithm.

A logical future direction is to use extra features in the existing methods,
as all the currently used features largely provide independent contributions.
For instance, geometry and temporal features can be added to the physical or texture-based approaches.
Alternatively, physical or texture features can be used to pre-select candidate shadow pixels
and feed them to geometry-based methods for shadow remodelling.
Lastly, even if there are considerable differences in the computational load across the various shadow detection methods,
all of them can be optimised to meet real-time requirements.
In particular, all the compared approaches are parallelisable at either the pixel or region level.

\section*{Acknowledgements}

NICTA is funded by the Australian Government
as represented by the {\it Department of Broadband, Communications and the Digital Economy},
as well as the Australian Research Council through the {\it ICT Centre of Excellence} program.
The authors thank the anonymous reviewers for useful suggestions.

\bibliographystyle{ieee}
\bibliography{references}

\end{document}